\documentclass{article}

\usepackage{arxiv}
\usepackage{graphicx}
\usepackage{amsmath,amssymb}
\usepackage{mathtools}
\usepackage{dsfont}
\usepackage{xspace}
\usepackage{epsfig}
\usepackage{algorithm}
\usepackage[noend]{algorithmic}
\usepackage{color}
\usepackage{url}
\usepackage{graphicx}
\usepackage{tikz}
\usepackage{wasysym}
\usepackage{booktabs}
\usepackage{longtable}
\usepackage{array}
\usepackage{multirow}
\usepackage{wrapfig}
\usepackage{float}
\usepackage{colortbl}
\usepackage{pdflscape}
\usepackage{tabu}
\usepackage{threeparttable}
\usepackage{threeparttablex}
\usepackage[normalem]{ulem}
\usepackage{makecell}
\usepackage{xcolor}
\usepackage{csquotes}

\newcommand{\wcity}{\ensuremath{\omega}}

\newcommand{\ttp}{{Traveling Thief Problem}}

\newcommand{\ttps}{TTP\xspace}
\newcommand{\tsp}{TSP\xspace}
\newcommand{\wttps}{W-TTP\xspace}
\newcommand{\nwdtsps}{W-TSP\xspace}

\newcommand{\EA}{$(1+1)$-EA\xspace}

\newcommand{\ignore}[1]{}

\begin{document}

\title{
Optimising Tours for the Weighted Traveling Salesperson Problem and the Traveling Thief Problem: A Structural Comparison of Solutions}
%
%
\author{
  Jakob Bossek \\
  Optimisation and Logistics\\
  The University of Adelaide\\
  Adelaide, Australia \\
  \texttt{jakob.bossek@adelaide.edu.au} \\
  \And
  Aneta Neumann \\
  Optimisation and Logistics\\
  The University of Adelaide\\
  Adelaide, Australia \\
  \texttt{aneta.neumann@adelaide.edu.au} \\
  \And
  Frank Neumann \\
  Optimisation and Logistics\\
  The University of Adelaide\\
  Adelaide, Australia \\
  \texttt{frank.neumann@adelaide.edu.au} \\
}

\maketitle              

\begin{abstract}
The Traveling Salesperson Problem (TSP) is one of the best-known combinatorial optimisation problems. However, many real-world problems are composed of several interacting components.
The Traveling Thief Problem (TTP) addresses such interactions by combining two combinatorial optimisation problems, namely the TSP and the Knapsack Problem (KP). Recently, a new problem called the node weight dependent Traveling Salesperson Problem (W-TSP) has been introduced where nodes have weights that influence the cost of the tour. In this paper, we compare W-TSP and TTP. We investigate the structure of the optimised tours for W-TSP and TTP and the impact of using each others fitness function.
Our experimental results suggest (1) that the W-TSP often can be solved better using the TTP fitness function and (2) final W-TSP and TTP solutions show different  distributions when compared with optimal TSP or weighted greedy solutions.

\end{abstract}


\section{Introduction}
\label{sec:introduction}

The Traveling Salesperson Problem (TSP) is one of the most prominent combinatorial optimisation problems and has been widely studied in the literature. It also serves as a basis for many more complex vehicle routing problems. Often real-world optimisation problems involve multiple interacting components that have to be optimised simultaneously. Moreover, due to the interactions the different silo problems can not be optimised separately in order to come up with an overall good solution~\cite{DBLP:books/sp/19/BonyadiM0019}.

The Traveling Thief Problem introduced in~\cite{Bonyadi2013TTP} is a multi-component problem that has recently gained significant attention in the evolutionary computation literature~\cite{Faulkner2015,ElYafrani:2016:PVS:2908812.2908847,ElYafrani2018231,DBLP:conf/gecco/WuP0N18,DBLP:conf/seal/Wu0PN17,DBLP:conf/gecco/WuPN16,DBLP:journals/soco/MeiLY16,Wagner2017ttpalgssel}. It combines the TSP and the classical Knapsack Problem by assigning items with profits and weights to the cities. The goal is to maximise the difference of profits of the collected items and the costs of a tour where the weights of items collected while visiting the cities increase the cost of moving from one city to the next one. More precisely, the weights of the items collected so far reduce the speed of the vehicle in a linear fashion and the cost of moving from city $i$ to city $j$ is determined by the current speed and the distance $d(i,j)$ of $i$ and $j$.
A wider range of benchmark instances have been introduced~\cite{DBLP:conf/gecco/PolyakovskiyB0MN14} and various competitions have been carried out at evolutionary computation conferences.

Understanding the interactions within the \ttps is difficult. If the given tour is fixed and only the remaining (still $\mathcal{NP}$-hard) packing problem has to be solved, then this can be done by dynamic programming and also approximation algorithms are available~\cite{DBLP:conf/algocloud/NeumannPSSW18}.
However, optimising the tour for the \ttps when the packing part is fixed seems to be significantly more difficult. In order to gain a better understanding on how node weights that influence the cost of a tour impact the optimisation, the node weight dependent Traveling Salesperson Problem (\nwdtsps) has been introduced recently~\cite{Bossek2020wtsp}. Here each node has a weight and the cost of going from city $i$ to city $j$ is their distance $d(i,j)$ times the weight of the nodes visited so far. For special cases approximation algorithms have been designed in \cite{Bossek2020wtsp} that establish a relation to the minimum latency problem~\cite{Blum_latency}. Furthermore, experimental investigations have been carried out to examine the impact of the node weights on the optimised salesperson tour.

With this paper, we continue this line of research and further bridge the gap in understanding the impact of node weights on salesperson tours. We examine and compare \ttps and \nwdtsps in a systematic study. We consider a variant to \ttps where the packing plan -- and in consequence the total profit -- is fixed and the goal is to minimise the cost of the weighted \ttps tour length. We call this problem \wttps. In our experimental investigations, we investigate instances where each item of a given \ttps benchmark is present with probability $p$.
Our study suggests, that with increasing $p$, i.e. increasing average number of nodes with strictly positive node weight, for the simple randomised search heuristic considered in this paper, it is advantageous to use the \wttps objective as a driver for the search process instead of the \nwdtsps objective in order to find good solutions for the \nwdtsps. Furthermore, we consider the difference in terms of the structure of solutions obtained using the different problem formulation.
In terms of structural similarity of \wttps and \nwdtsps solutions produced by our simple heuristic, good \nwdtsps on average show higher similarity with the solutions obtained by a naive weighted greedy approach (WGR) then this is the case for \wttps solutions with respect to a similarity measure based on the inversion number. In contrast, good \wttps solutions on average share more edges with optimal TSP solutions. We hope that in future such findings can be leveraged to develop more sophisticated heuristic search algorithms for both the \nwdtsps and the \ttp.

The paper is structured as follows. We introduce the problems examined in this paper in Section~\ref{sec:problem}, and we carry out our experimental investigations in Section~\ref{sec:experimental_setup}. Afterwards, in Section~\ref{sec:solution_quality}, we investigate the relation of solutions among the two problems in terms of objective value ratios. In Section~\ref{sec:similarity_analysis} we perform a structural similarity analysis of solutions with optimal TSP tours and weighted greedy solutions. We finish with some concluding remarks and avenues for future research.


\section{Problem Formulation}
\label{sec:problem}

The classical Traveling Salesperson problem is one of the most studied $\mathcal{NP}$-hard combinatorial optimisation problems. Given a set of $n$ cities $V =\{1, \ldots, n\}$ and distances $d(i,j)$ between them, the goal is to find a permutation $\pi$ which minimizes the tour length given by
$$
\text{TSP}(\pi) = d(\pi_n, \pi_{1})+ \sum_{i=1}^{n-1}  d(\pi_i, \pi_{i+1}).
$$

Motivated by the TTP, we study variants of this problem where node weights influence the cost of a tour.

\subsection{The Traveling Thief Problem}

The Travelling Thief Problem (\ttps) was first introduced in~\cite{Bonyadi2013TTP}. Given is a set of $n$ cities $V =\{1, \ldots, n\}$ with pairwise distances $d(i,j)$ between them and a set $E_i=\{e_{i1}, \ldots, e_{im_i}\}$ of $m_i=|E_i|$ items at city $i$, $1 \leq i \leq n$. We denote by $E = \cup E_i$ the overall set of items.
There is a profit $p\colon E \rightarrow \mathds{R}^+$ and weight function $w\colon E \rightarrow \mathds{R}^+$ on the items and knapsack capacity $C$ which limits the total weight of a selection of items.

The goal in the \ttps is to find a tour $\pi=(\pi_1, \ldots, \pi_n)$ and a packing plan $x= \left(x_{11},\ldots,x_{nm_n}\right)$ such that their combination $\pi$ and $x$ maximises the sum of the profits minus the travel cost associated with $\pi$ and $x$. Note that in the classical TTP, there is usually no item available at city $1$.

We indicate by a bitstring $x=\left(x_{11}, \ldots, x_{nm_n})\right)  \in \{0,1\}^m$, where $m= \sum_{i=1}^n m_i$, the items present in a problem instance. Item $e_{ij}$ is present iff $x_{ij}=1$ holds.

We denote by
$$w(\pi_i,x) = \sum_{k=1}^{m_{\pi_i}} w(e_{\pi_i k}) x_{\pi_ik}$$
the weight of the items taken in city $\pi_i$ with packing plan $x$. The number of present items at city $\pi_i$ is $$\eta(\pi_i)=\sum_{k=1}^{m_{\pi_i}} x_{\pi_ik}.$$
In our experiments, we consider the case where all cities have the same number of items and use the notion IPN for \emph{items per node}.

Let $\wcity(i)=\sum_{j=1}^i w(\pi_j,x)$ be the sum of the weights of the cities in permutation $\pi$ up to the $i$th city.
The cost of a tour is given by the time the vehicle takes to complete the tour. Here the weight of the items present when going from city $i$ to city $j$ depends on the distance $d(i,j)$ and the speed $\upsilon \in [\upsilon_{\min}, \upsilon_{\max}]$, where $\upsilon_{\min}$ is the minimum speed and $\upsilon_{\max}$ is the maximum speed of the vehicle. The tour has to start and city $1$ and therefore $\pi_1=1$ is required.

The goal in the standard formulation of TTP is to maximize
\begin{align*}
\text{\ttps}(\pi,x) = & \displaystyle \sum_{e\in E} p(e) x_{e} -R \left( \frac{d(\pi_n, \pi_{1})}{\upsilon_{max}-\nu \wcity(n)} + \displaystyle\sum_{i = 1}^{n-1}\frac{d(\pi_i, \pi_{i+1})}{\upsilon_{max}-\nu \wcity(i)} \right) \label{wttp}
\end{align*}
where $\sum_{e\in E} p(e)$ is the sum over all packed items’ profits, $\nu = \left(\upsilon_{max}-\upsilon_{min}\right)/C$ is a constant value defined by the input and $R$ is a constant called the renting rate.

We assume that the packing plan is fixed $x$ for a given instance.
If $x$ is fixed then the profits and the weights at the cities are completely determined.
 We ignore the profit part and the renting rate as both are constant and do not have any impact on the order of solutions with respect to the fitness function \ttps.
In our study, we investigate the following cost function which depends on the weights of the items determined by $x$ and the chosen permutation $\pi$:
\begin{align*}
\text{\wttps}(\pi,x) = & \displaystyle \left( \frac{d(\pi_n, \pi_{1})}{\upsilon_{max}-\nu \wcity(n)} + \displaystyle\sum_{i = 1}^{n-1}\frac{d(\pi_i, \pi_{i+1})}{\upsilon_{max}-\nu \wcity(i)} \right) 
\end{align*}

We call the problem of finding a tour which minimizes this goal function the weighted \ttps-problem (\wttps).

\subsection{The Node Weight Dependent TSP}

We also consider the node weight dependent TSP problem (\nwdtsps) recently introduced in~\cite{Bossek2020wtsp}.
In addition to the input of the TSP, we have a set of possible items $E_i$ available at each city $i$. Following the notation for \wttps, we indicate by a bitstring $x \in \{0,1\}^m$ whether an item $e_{ij}$ is present.

Given a set  of $n$ cities $V=\{1, \ldots, n\}$ with distances $d(i,j)$ between the cities  and a weight function $w \colon E \rightarrow \mathds{R}^+$ on the set of items, the goal is to find a permutation $\pi$ that minimizes the weighted TSP cost. The tour has to start and city $1$ and therefore $\pi_1=1$ is required.
We denote by
$$w(\pi_i,x) = \sum_{k=1}^{m_{\pi_i}} w(e_{\pi_i k}) x_{\pi_ik}$$

the weight of the items presents at city $\pi_i$.
The fitness of a given tour $\pi$ and a given set of present items indicated by $x$ is given as
\[
\text{\nwdtsps}(\pi,x)=d(\pi_n, \pi_{1})\left(\sum_{j=1}^n w(\pi_j,x) \right)+ \sum_{i=1}^{n-1}  d(\pi_i, \pi_{i+1})\left(\sum_{j=1}^i w(\pi_j,x) \right).
\]

Note, that the standard TSP is the special case where $w(\pi_1)=1$ and $w(\pi_i)=0$, $2 \leq i \leq n$.

Our fitness function definitions for \wttps and \nwdtsps work with a set of present items which can also be defined in terms of the input items without using the bitstring $x$.
We use the notation of present items indicated by $x$ as we will use TTP benchmarks where different subsets of items of a given TTP instance have to be collected in the computed tour.

\subsection{Problem Comparison}

The \tsp, \wttps, and \nwdtsps place different emphasize on the weight of nodes. The TSP can be considered as the special case of \nwdtsps where only the first node receives a weight of $1$. Furthermore, TSP is a special case of the tour optimisation variant of \ttps where no item is collected, and the vehicle always travels at maximum speed $v_{\max}$.
\nwdtsps allows for a very drastic and high weightening of distance costs as the weights are collected during the route and each distance is multiplied with the weight of the cities visited. \ttps in more limited in terms of the impact of the weightening as the weight of the items reduces the speed from $v_{\max}$ to $v_{\min}$ in a linear fashion. Using the interval $[v_{\min}, v_{\max}]$ for the speed also ensures that the weighted distance for going from city $i$ to $j$ is always in the interval
$[d(i,j)/v_{\max}, d(i,j)/v_{\min}]$ where as in the case of \nwdtsps this can be in the range $[0,W\cdot d(i,j)]$ where $W$ is the total weight amount all cities.


\begin{figure}[ht]
    \centering
    \includegraphics[width=\textwidth]{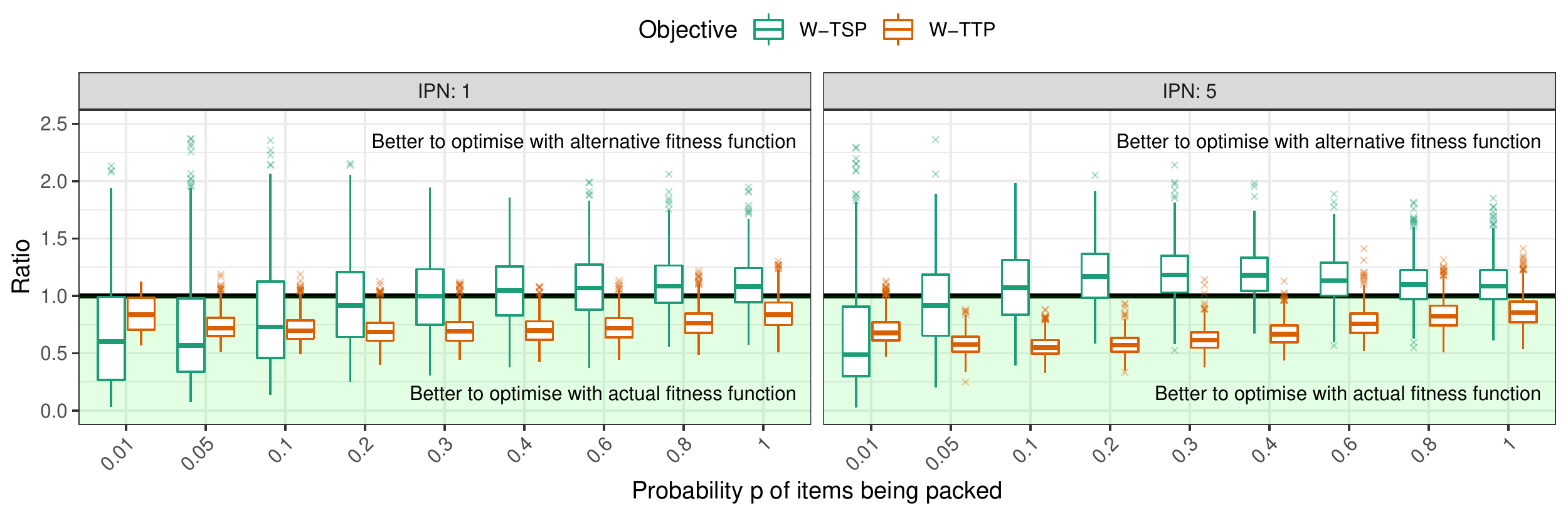}
    \caption{Distribution of objective value ratios of final tours. Ratios are calculated by the following rule: if \nwdtsps is to be minimised we divided the \nwdtsps tour-length obtained by optimising with the actual \nwdtsps driver with the \nwdtsps tour-length of the solution calculated when the algorithm is run with the \wttps driver instead. Ratios for \wttps optimisation are calculated analogously. Ratios below zero indicate a benefit for the actual objective function.}
    \label{fig:boxplots_all_ratios}
\end{figure}

\section{Experimental Setup}
\label{sec:experimental_setup}

The focus of this paper is on understanding interactions between solutions for the \wttps and the recently introduced \nwdtsps. To study these effects, we consider a subset of instances from the \ttps 2017 CEC Competition\footnote{ \url{https://cs.adelaide.edu.au/~optlog/TTP2017Comp/}.} for our experiments~\cite{DBLP:conf/gecco/PolyakovskiyB0MN14}. We choose all instances which are based on the following classical TSPlib~\cite{Reinelt91tsplib} instances: a280, berlin52, ch130, ch150, eil101, eil51, eil76, kroA100, kroC100, kroD100, lin105, pcb442, pr1002, pr2392, pr76, rd100, st70. Therein, all three weight/profit classes are covered: bounded strongly correlated (bsc), uniform similar weights (usw) and uncorrelated (u). Furthermore, the number of items per node (IPN) is either one or five. In total our benchmark set contains $102$ instances. The subset is a cross-section of the \ttps benchmark set with instances of few nodes up to instances with several thousand nodes. In addition, optimal tours for the classical TSP are known for these instances. This will be of essential for structural similarity analysis in Section~\ref{sec:similarity_analysis}.
Recall that in our setup the packing plan is initially fixed and so are the weights at the nodes; no changes to the packing are made in the course of optimisation. To account for the stochasticity in the packing and the influence of the fraction of active items, for each instance and each $p \in \{0.01, 0.05, 0.1, 0.2, 0.3, 0.4, 0.6, 0.8, 1.0\}$ we generated $31$ random packings from a $\text{Bin}(m, p)$-distribution where $m$ is the number of items of the \ttps instance at hand, i.e. each items is packed with probability $p$ and not packed with inverse probability $(1-p)$. In order to make all generated packings feasible, we set the knapsack capacity $C$ to the sum of \underline{all} item weights (not just the packed ones).\footnote{Note that this step is relevant for the \wttps only; the \nwdtsps objective function does not cope with a knapsack limit.} Note that this choice for the knapsack capacity allows us to explore different degrees of filling of the vehicle. In consequence a transition from the classical TSP ($p$ close to zero) and the TTP with a fully loaded vehicle ($p$ close to one) is possible.

We consider the classical \EA with inversion mutation on permutations. Preliminary benchmarking with swap and insertion mutation showed its superiority; this confirms the experimental results in~\cite{Bossek2020wtsp} on the \nwdtsps. We urge the reader to carefully read the following sentences as they convey a crucial aspect of our study: we run \EA with either the \wttps or the \nwdtsps for driving the evolutionary search process (EA driver). In addition, the best so far solution in every iteration and in particular the final best solution is evaluated with both \wttps and \nwdtsps resulting in four different relevant combinations.

\EA is applied each one time on each instance and each of the 31 associated packings plans. Note, that we do not perform additional independent runs for each fixed packing plan. Instead, the 31 runs already account for the stochasticity.
Our implementation and data is available in a public GitHub repository.\footnote{GitHub repository: \url{http://github.com/jakobbossek/ttp}}


\section{Comparison in Terms of Solution Quality}
\label{sec:solution_quality}

We first approach the following research question: is it beneficial to use each others fitness function for optimisation purposes? More precisely, if we aim to optimise the \wttps (\nwdtsps), should we use the actual objective function as EA driver or is it of benefit to use the \nwdtsps (\wttps) objective function instead? One might argue that it certainly makes no sense to use another fitness function as a surrogate. However, our results prove this assumption wrong in many cases.
Figure~\ref{fig:boxplots_all_ratios} show the distribution of objective value ratios across all runs on all considered instances separated by the instance property IPN and the packing probability $p$. The ratios are to be interpreted as follows: when the objective is \nwdtsps we divide the \nwdtsps objective value of the final solution determined with the \nwdtsps-driver by the \nwdtsps objective value of the final solution obtained by optimising with the \wttps-driver and vice versa. Since both objectives are to be minimised a ratio below $1.0$ indicates that it is advantageous to use the actual objective function to guide the EA; the result one would expect. Returning to Figure~\ref{fig:boxplots_all_ratios} we actually see that this assumption does not always hold true; at least in one direction. The data shows that it is consistently advisable to use the \wttps objective function to optimise the \wttps. However, a closer look shows that the \wttps-related box-plots show a characteristic U-shape with peaks in the area of $p \approx 0.5$. In contrast, with \nwdtsps being in the focus of optimisation we observe a very different pattern. Here, with $p \to 1$, the median ratio increases. The median surpasses $1.0$ for the first time at a level of $p=0.4$ with one item per node and $p=0.1$ for $\text{IPN}=5$. Our assumption is that for $\text{IPN}=1$ and given $p \in [0,1]$ in expectation $np$ nodes have a strictly positive weight. In contrast, if there are multiple items per node, due to independence of the item activation in the packing plan generation, in each node $m_{i}p$ are expected to be active. Hence, in expectation, there will be more nodes with strictly positive weight assigned in this setting. Either case it seems as with increasing $p$ oftentimes the \wttps-driver leads to better \nwdtsps tours. The results suggest that using the \nwdtsps objective produces large basins of attraction for qualitatively bad local optima.
\begin{figure}[t]
    \centering
    \includegraphics[width=\textwidth]{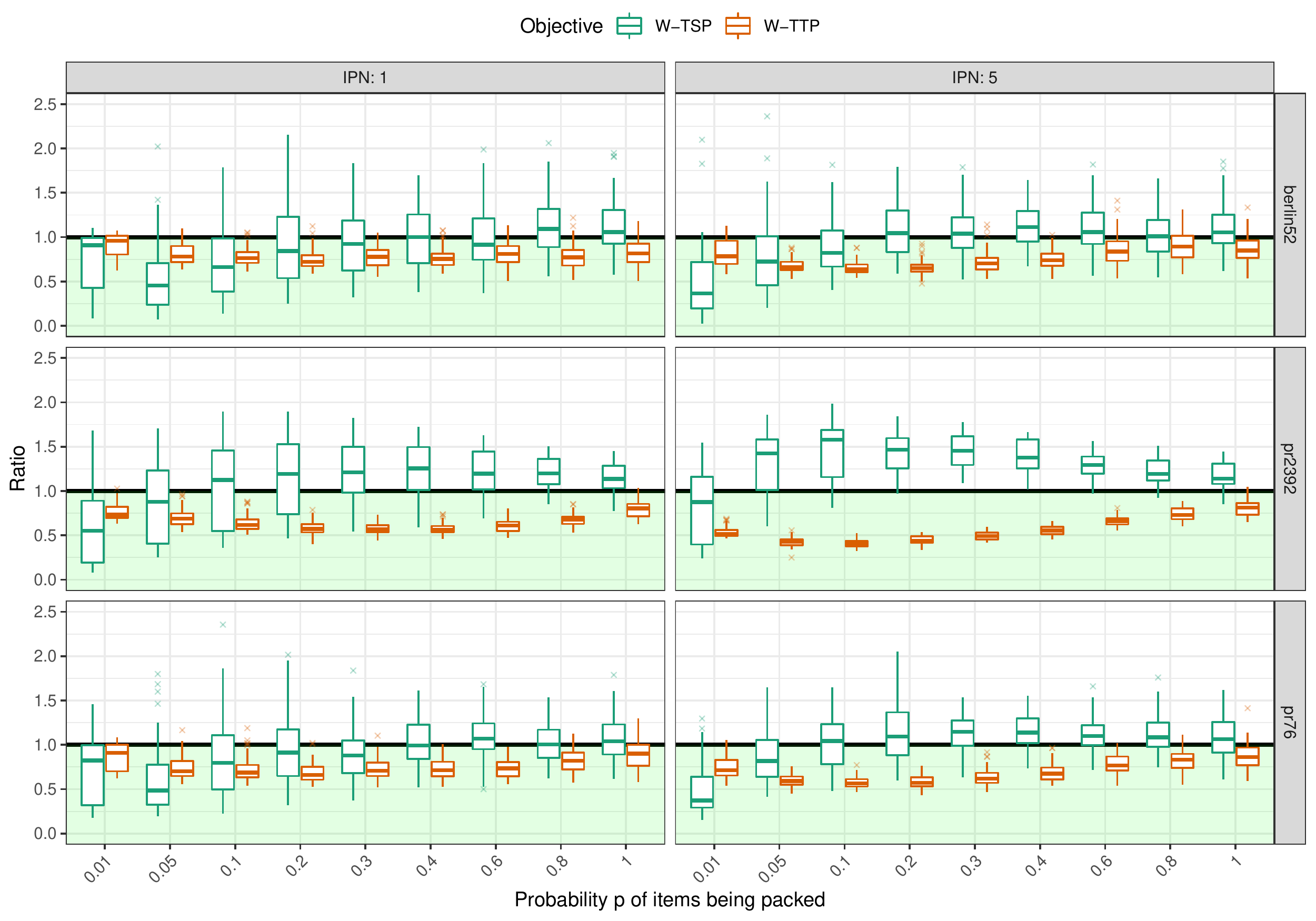}
    \caption{More fine-grained objective ratios for three representative instances (rows) and different item counts (columns).}
    \label{fig:boxplots_selected_ratios_fine-grained}
\end{figure}
Figure~\ref{fig:boxplots_selected_ratios_fine-grained} shows a less aggregated view. Here, the ratios are shown for three representative instances from the benchmark set (still aggregated across weight/profit types bsc, usw and u since the type does not reveal any different patterns). Here, in particular the largest pr2392-based instances with $n=2\,392$ nodes stands out from the crowd: here the aforementioned U-shape observed for the \wttps is inverse for the \nwdtsps at least for $\text{IPN}=5$. For this particular instance the difference between median ratios is highest and using the \wttps EA-driver for moderate $p$ leads to median quality gains of $\geq 1.5$ which is massive.

\begin{figure}[tbp]
    \centering
    \includegraphics[width=0.325\textwidth]{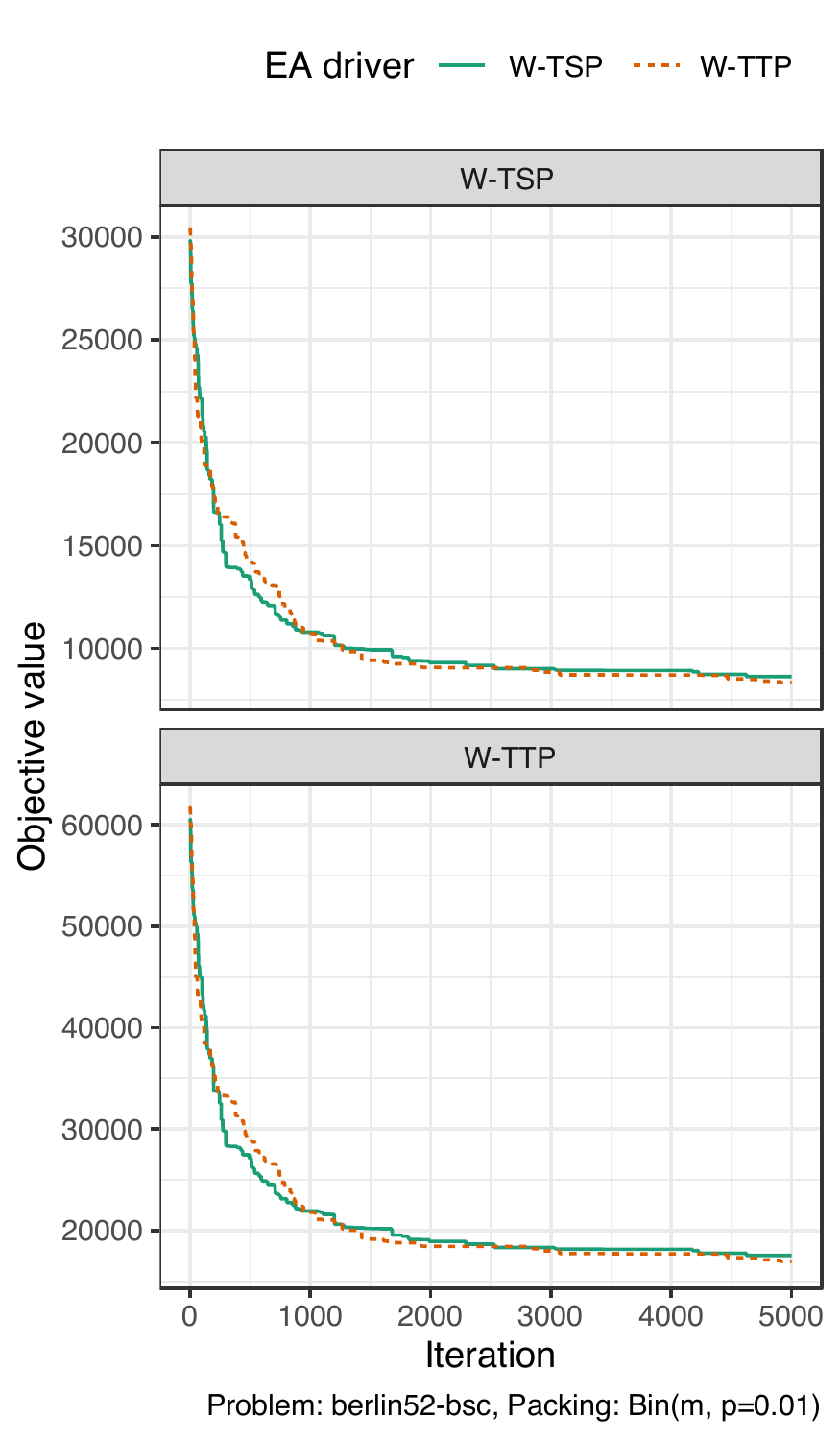}
    \includegraphics[width=0.325\textwidth]{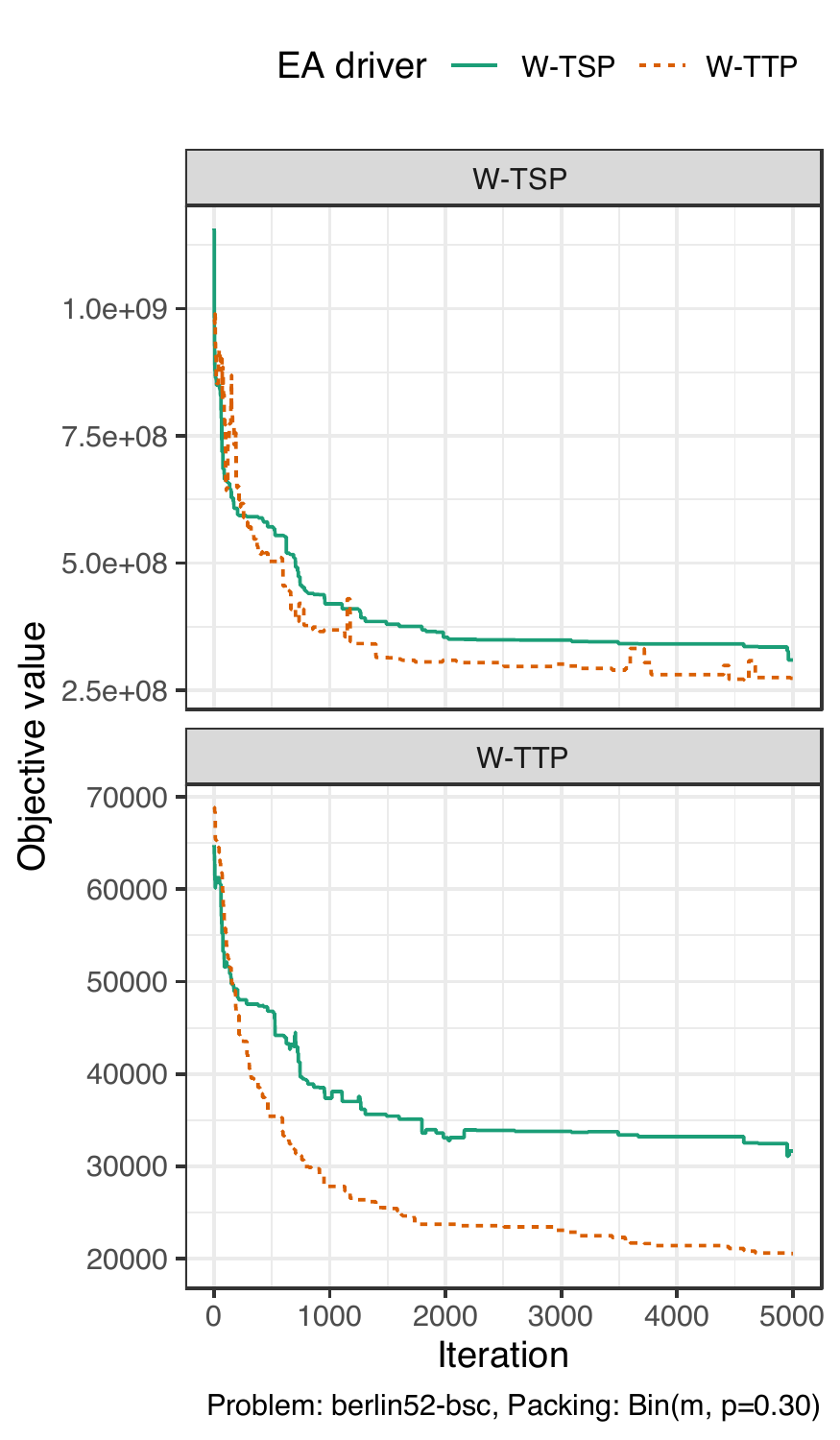}
    \includegraphics[width=0.325\textwidth]{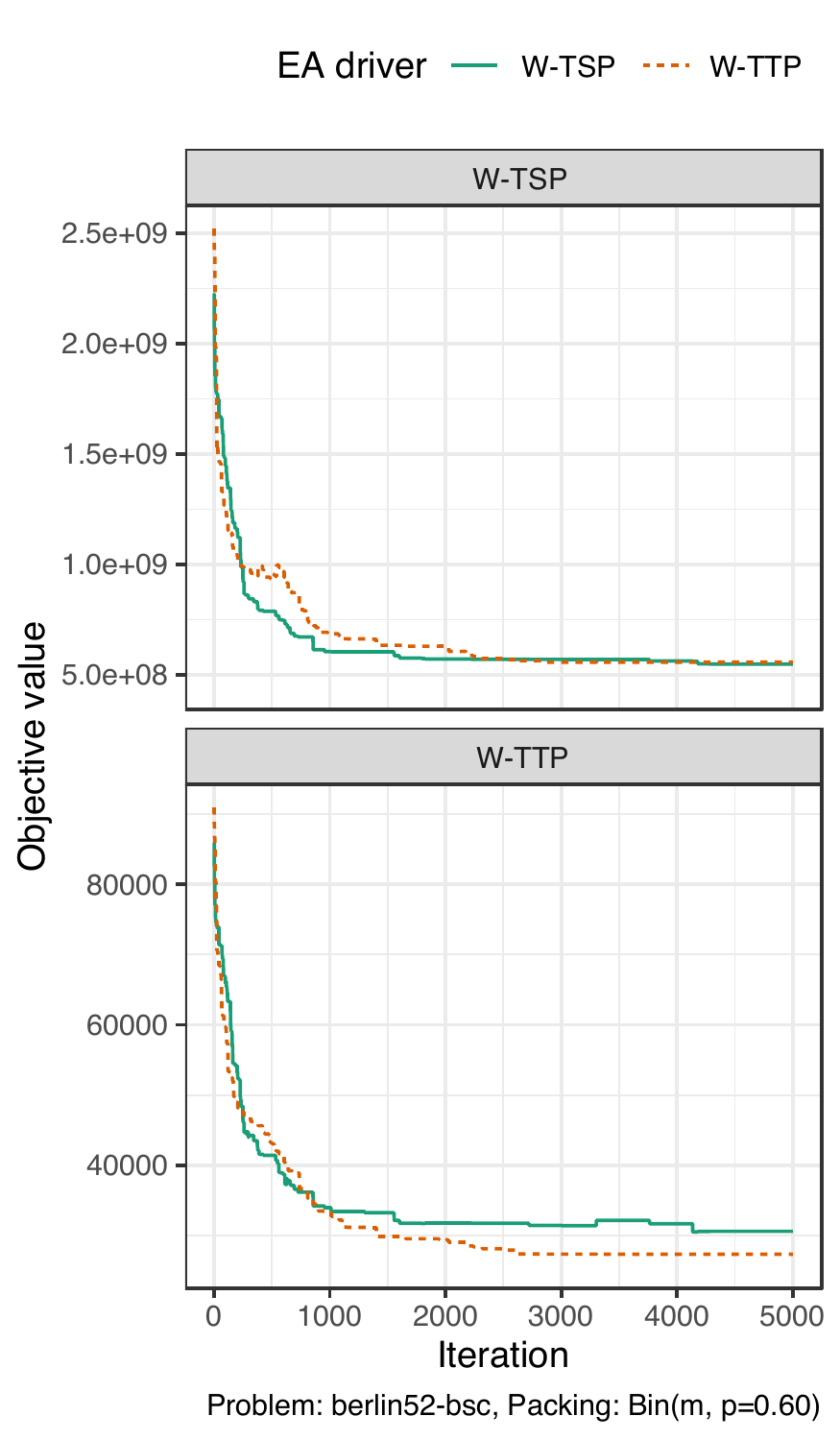}

    \vspace{0.75cm}
    \includegraphics[width=0.325\textwidth, trim=0 0 0 30pt, clip]{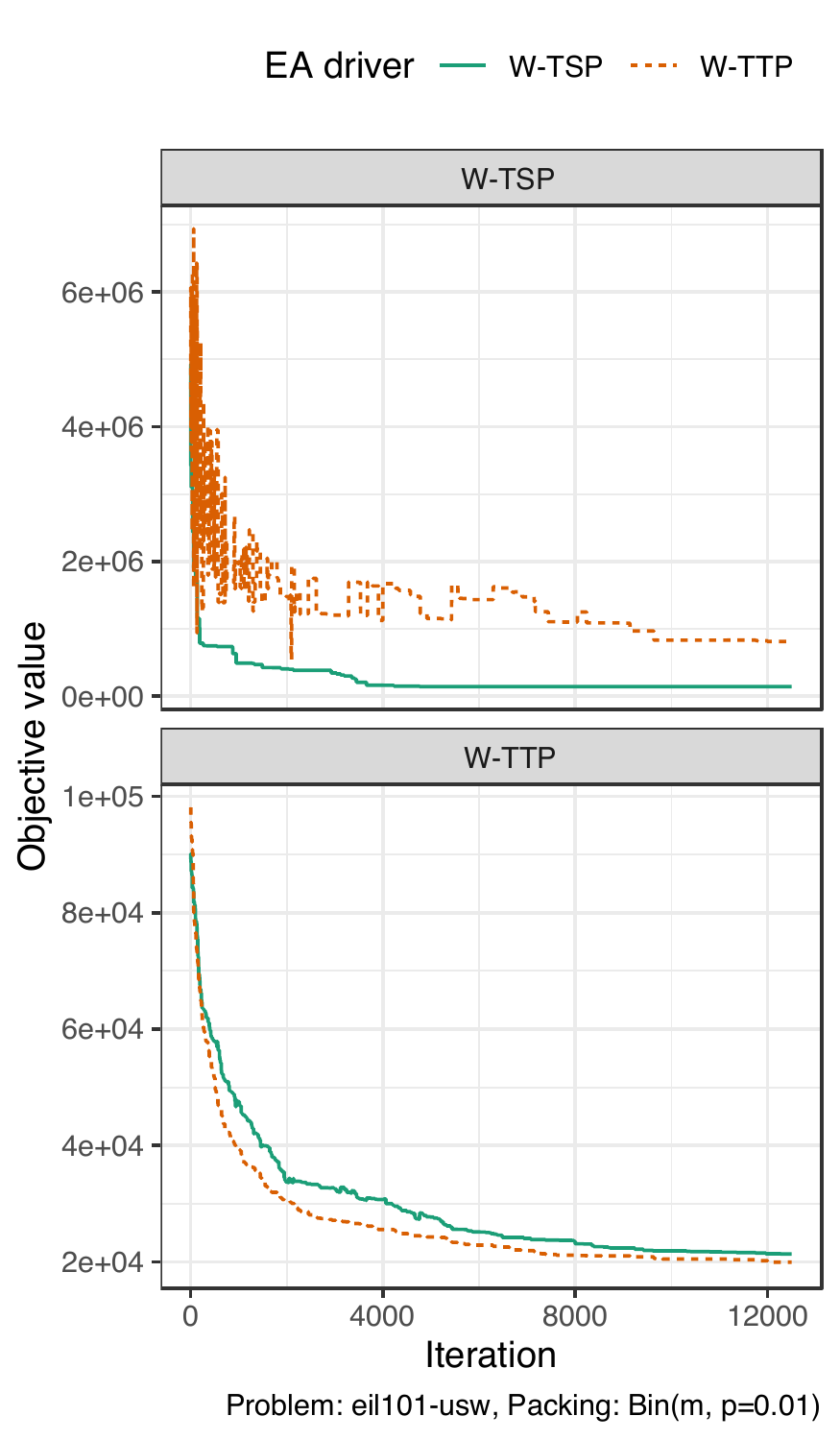}
    \includegraphics[width=0.325\textwidth, trim=0 0 0 30pt, clip]{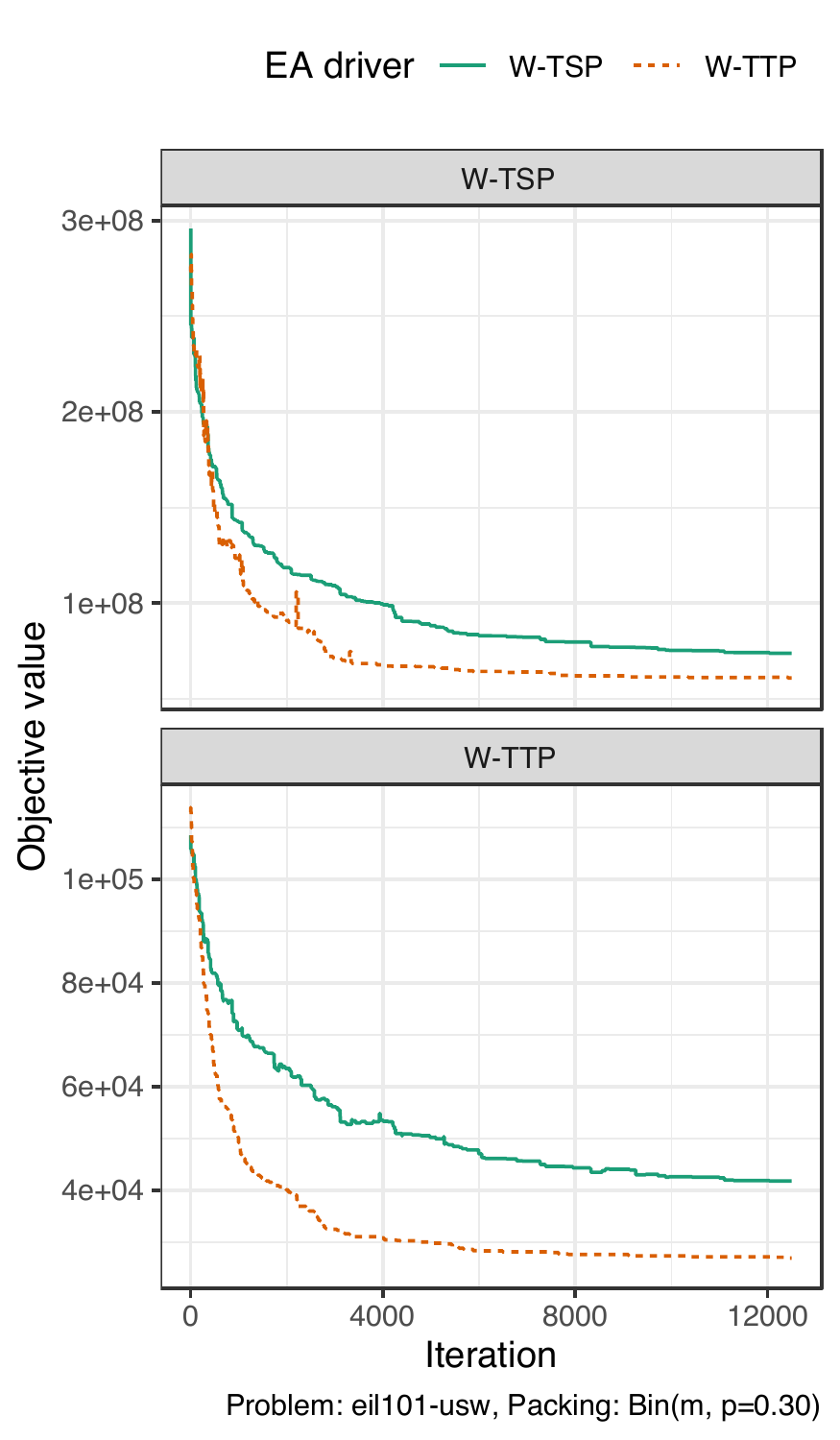}
    \includegraphics[width=0.325\textwidth, trim=0 0 0 30pt, clip]{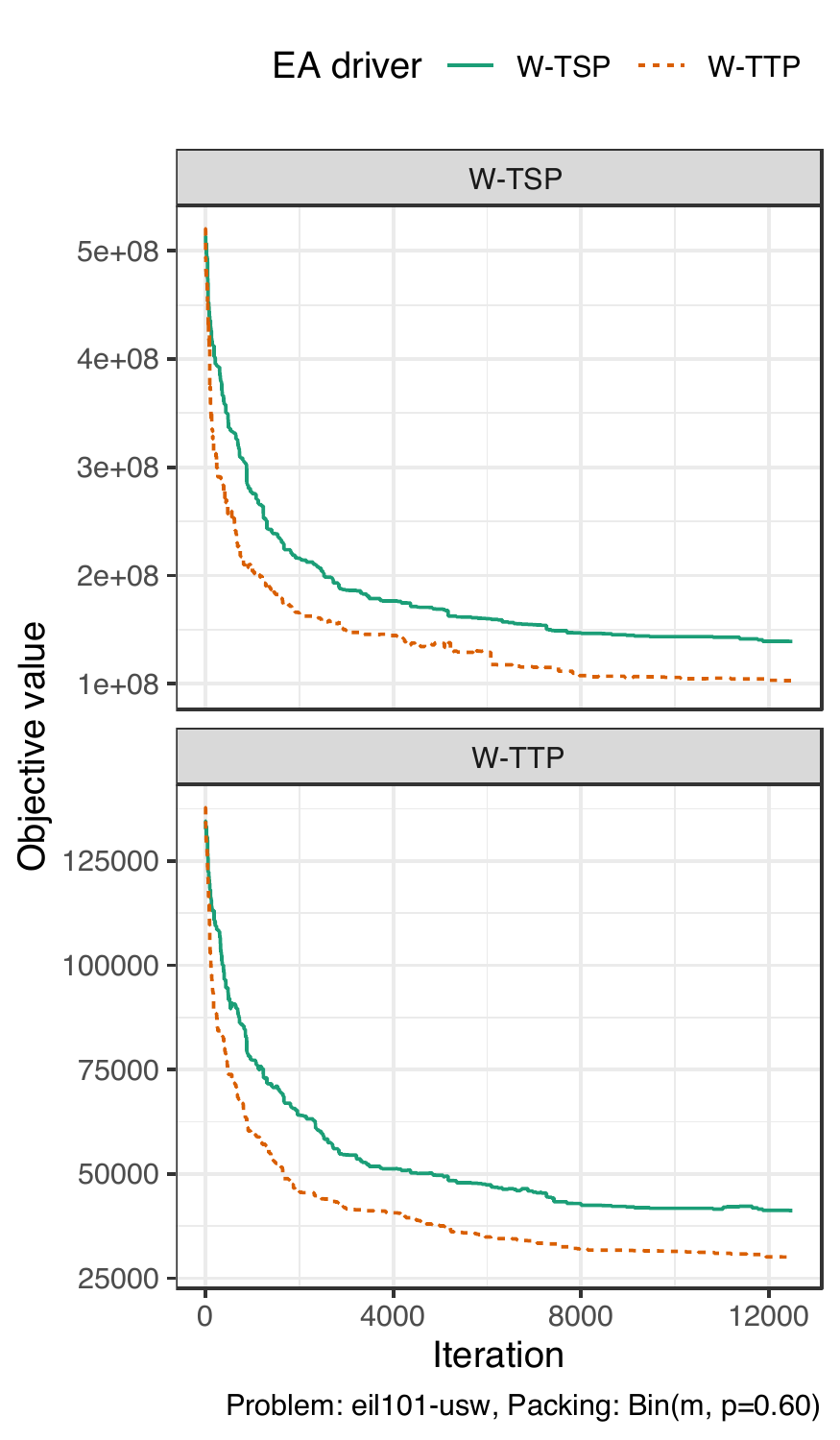}

    \caption{Exemplary trajectories for instance berlin52 (two top rows) and eil101 (two bottom rows) with bounded strongly correlated weights and 5 items per node. The EA was run with both \nwdtsps and \wttps as driver (indicated by color and line type). Likewise, incumbent solutions were evaluated with both objective functions (\nwdtsps in top and \wttps in bottom row).}
    \label{fig:trajectories}
\end{figure}
Figure~\ref{fig:trajectories} visualises the trajectories / development of incumbent solutions for two representative instances. In particular for $p=0.3$ (second column) we see that for these particular runs in fact the final \nwdtsps objective is better when the EA driver is \wttps. Moreover, occasional decrease in fitness values can be observed even though the general optimisation goal is still purchased.

\begin{figure}[ht]
    \centering
    \includegraphics[width=\textwidth, trim = 20pt 10pt 20pt 20pt, clip]{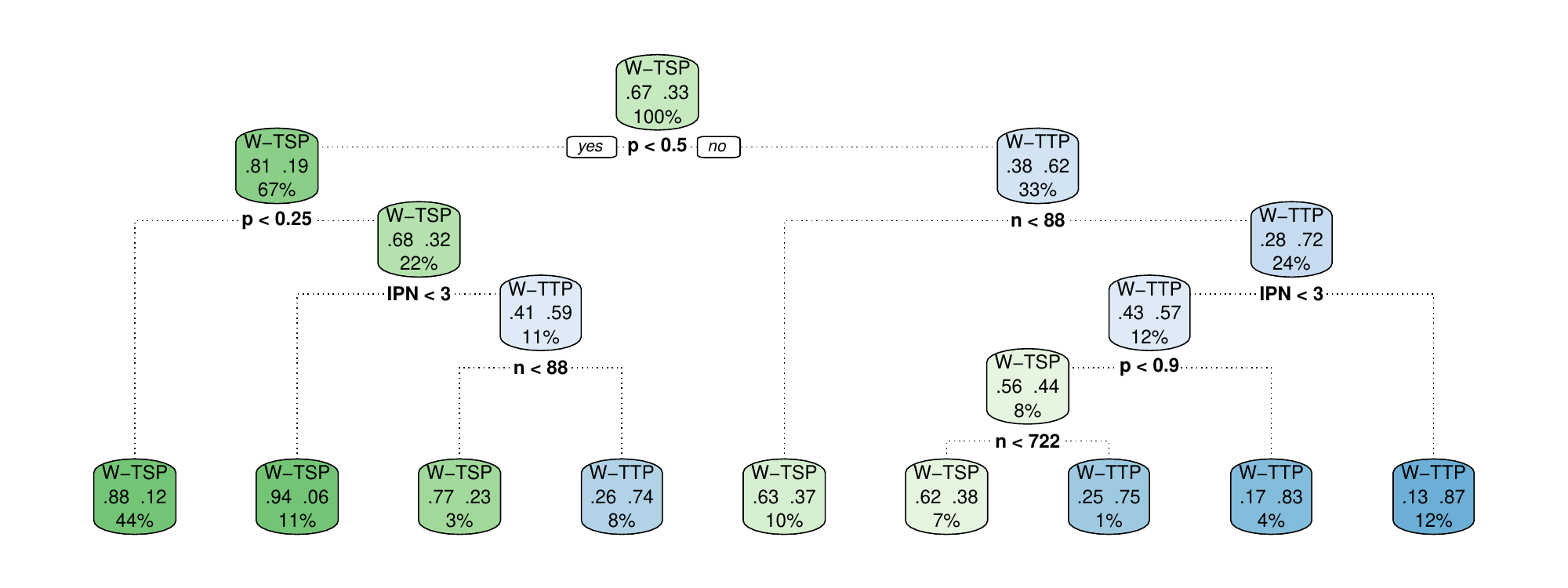}
    \vspace{-0.2cm}
    \caption{Decision tree for the machine learning task of determining which objective function should be used in order to optimise the \nwdtsps. Within the splits $p$ is the probability of items being active, $n$ is the number of nodes and IPN is the number of items per node. Values within the nodes indicate the majority decision (top), the fraction of data processed by the left/right branch respectively (center) and the percentage of overall data points processed at that node.}
    \label{fig:decision_tree}
\end{figure}

In order to make sense out the data we trained a simple decision tree to decide which EA-driver to use in order to solve the \nwdtsps. Since the \wttps is best solved by adopting the \wttps driver (beside few outliers) we did not perform this step for the other direction. Our goal was a simple binary classification task. I.e. the target is to decide which EA-driver is preferable while predictor variables are the instance size $n$, the IPN value and the probability $p$. We used 10-fold cross-validation and the R-package \texttt{rpart}~\cite{Rrpart} interfaced by package \texttt{mlr}~\cite{Rmlr} to train the model and access its performance. The cross-validation results report a mean miss-classification test error of $18.5\%$ and thus an accuracy of $81.5\%$ in predicting the best EA-driver. This is not overwhelming, though admittedly higher than tossing a coin. The final decision tree is depicted in Figure~\ref{fig:decision_tree}. The splits used by the model, i.e. decisions made when we follow the nodes from the root down to leaf level, very much reflect our previous observations where the \wttps driver is advantageous for larger $p$ and $\text{IPN}>1$.


\section{Structural Similarity Analysis of Solutions}
\label{sec:similarity_analysis}

In the following we conduct a similarity analysis of solutions. To be more concise we investigate the similarity of final \wttps and \nwdtsps solutions calculated in our study with two types of permutations: (1) optimal TSP solutions for the underlying TSP instance and (2) tours calculated by a greedy algorithm which favors visiting \enquote{heavy} nodes, i.e. nodes of high weight, later in the tour. In a nutshell the algorithm termed \emph{weighted greedy} (WGR) works as follows. In a first step nodes are  sorted in ascending order of their node weight. The second step is about tour construction. Here, nodes are visited in ascending order of node weight. In case of ties, i.e. several nodes with the same node weight, these nodes are visited following the nearest neighbor heuristic~\cite{lawler1985travelling}. This construction method can be seen as a naive approach to solve the \wttps or \nwdtsps respectively where one might assume that nodes with a high weight loading should be visited later on even if this requires to take some long distance edges beforehand. Note that the optimal TSP tours and WGR tours pose two extremes: the TSP tour is focused on the distances only  neglecting node weights completely. In contrast, WGRs' focus, though not able to guarantee optimality, is mainly on late heavy node placement in the tour.

For the purpose of measuring similarity we use two metrics for the comparison of two tours (permutations) $\pi^{1}$ and $\pi^{2}$. The first is termed \emph{common edges} (CE) and is defined as the proportion of edges shared by both tours. The second metric is based on the mathematical term of \emph{inversion} which -- in the classical sense -- is a measure of the sortedness of a sequence: for a permutation $\pi$, if $1 \leq i < j \leq n$ and $\pi_i > \pi_j$ the pair $(i,j)$ is called an inversion~\cite{Vitter1991AverageCaseAnalysis}. The total count of inversions IN$(\pi^{1},\pi^{2})$ is termed the \emph{inversion number} which is at most $n(n-1)/2$ with higher values indicating stronger dissimilarity with respect to sortedness.
In our setting though we are given two permutations $\pi^{1},\pi^2$ and we call $(i,j)$ an inversion, if node $i$ is visited before (after) node $j$ in $\pi^{1}$ and after (before) $j$ in $\pi^{2}$. In order to obtain a normalised similarity version we define our second measure as follows:
$$
\text{INV}(\pi^{1}, \pi^{2}) := 1 - \left(\frac{2 \cdot \text{IN}(\pi^{1},\pi^{2})}{n(n-1)}\right) \in [0, 1].
$$

\begin{figure}[t]
    \centering
    \includegraphics[width=\textwidth]{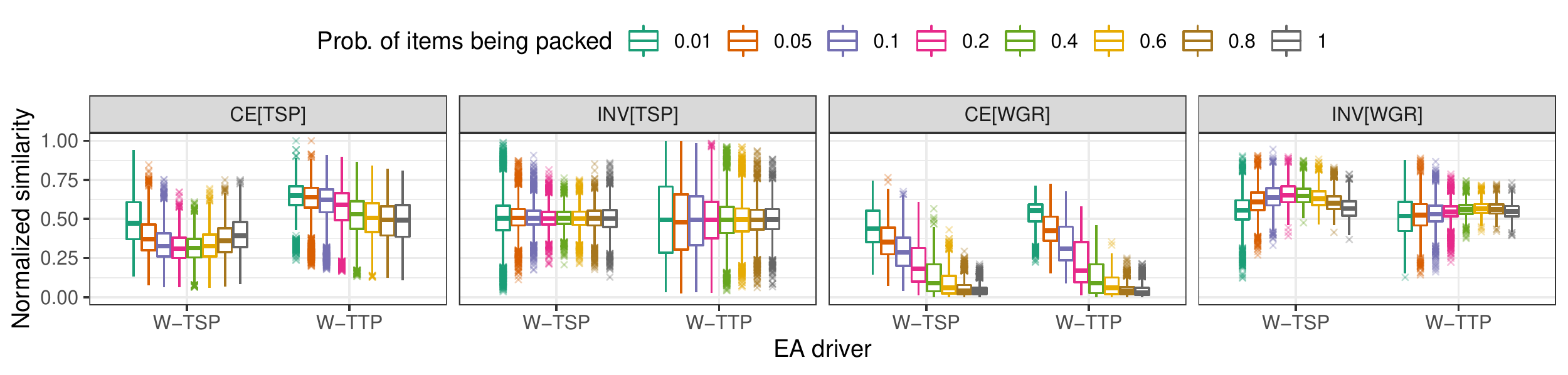}
    \vspace{-0.6cm}
    \caption{Distribution of similarity of all final
    \nwdtsps and \wttps solutions calculated in our experimental study. We calculate the similarity to the optimal tour for the classical TSP and the weighted greedy tours (WGR) respectively.}
    \label{fig:boxplots_similarity_aggregated_across_all_WGR}
\end{figure}

We want to stress that with a simple heuristic like the \EA it is unlikely to get optimal solutions to our problems. In consequence, the following observations are based on sub-optimal approximations to the \wttps and \nwdtsps respectively. Nevertheless, we believe that that our insights are valuable first steps towards a better understanding of tour composition.

Figure~\ref{fig:boxplots_similarity_aggregated_across_all_WGR} shows the distribution of the similarity of \nwdtsps and \wttps solutions with optimal TSP tours and WGR tours by means of the two measures CE and INV throughout the whole benchmark set. For ease of reference, we denote the similarity with CE[TSP], CE[WGR], INV[TSP] and INV[WGR]. Regarding CE[TSP]-similarity we observe a U-shape with increasing probability $p$ for \nwdtsps. The box-plots for \wttps however show a clear downward trend, i.e. the more items have to be collected by the thief, the less similar the tour gets to the TSP. Nevertheless, for both \wttps and \nwdtsps the median similarity is larger than $25\%$ for all values of $p$ and even above $50\%$ for the \wttps. Compared with this for both considered optimisation problems the CE[WGR]-similarity strongly decreases with increasing $p$. Here, median values close to $0\%$ with low variance are reached if on average at least $60\%$ of the items are active. The CE-measure is plain simple and kind of binary in the sense that an edge is either shared or not. However, even if the number of shared edges approaches zero the INV-similarity can show different patterns as it measures the number of swaps needed  to transform one tour into another. In fact, median INV[WGR]-values are $>50\%$ for all considered settings and both \nwdtsps and \wttps. Moreover, with increasing $p$ there is trend towards a narrowed outlier distribution, i.e. outliers are less frequent indicating a lower total range of similarity values. In addition, for the \nwdtsps we observe an inverted U-shape with its median peak at about $p=0.2$. This suggests that for the \nwdtsps and a relatively low number of active items it is in fact advisable to place these heavy nodes in the end of the permutation.
\begin{figure}[t]
    \centering
    \includegraphics[width=\textwidth, trim=0 5pt 0 11pt, clip]{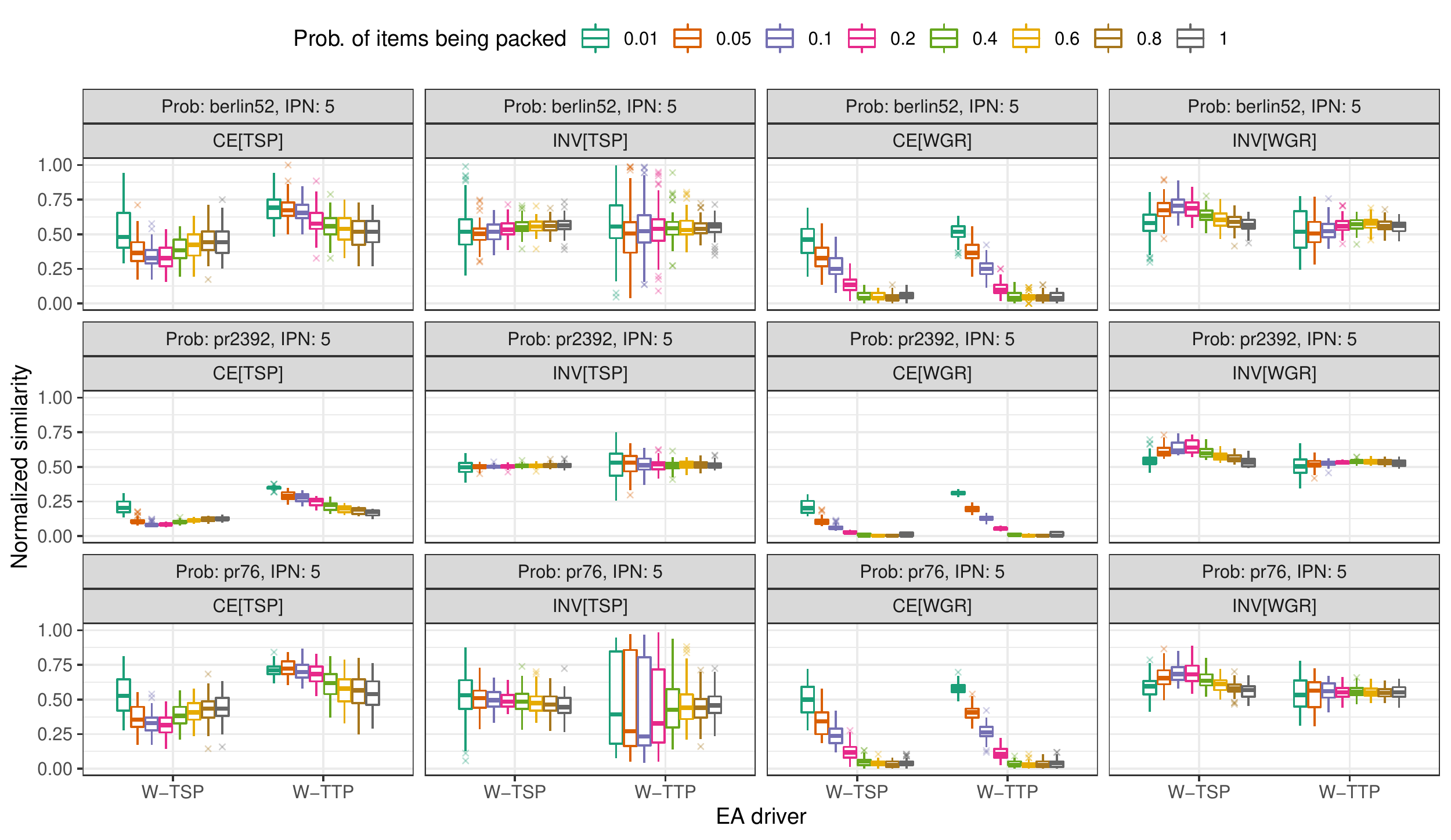}
    \vspace{-0.3cm}
    \caption{Distribution of similarity of all
    \nwdtsps and \ttps solutions calculated for instances of type berlin52 (top row), pr2392 (middle row) and pr75 (bottom row) to the respective optimal TSP tours and weighted greedy tours (WGR).}
    \vspace*{-0.5cm}
    \label{fig:boxplots_similarity_aggregated_for_examplary_instances}
\end{figure}
All observations made so far are valid for all considered instances and IPN values (see Figure~\ref{fig:boxplots_similarity_aggregated_for_examplary_instances} for a less aggregated view for three representative instances). We clearly observe the same patterns even though the actual similarity values can differ substantially (cf. the CE[TSP]-similarity in Figure~\ref{fig:boxplots_similarity_aggregated_for_examplary_instances}). In particular pr2392-based instances stand out. This is partly explained by its size ($2\,392$ nodes) which is much bigger than the majority of our benchmark instances and the fact that we use a very simple heuristic. Therefore, our \nwdtsps and \wttps solutions for those instances are likely far away from optimal.

Coming back to the actual measures: the only measure which shows strong variance throughout the instance set is INV[TSP]. This observation can be visually derived from Figure~\ref{fig:boxplots_similarity_aggregated_across_all_WGR} where we see many partly extreme outliers and is backed up by the representative more fine-grained plots in Figure~\ref{fig:boxplots_similarity_aggregated_for_examplary_instances}. The strong variance is even more pronounced for the \wttps solutions. To be honest, at this point we have no clear explanation to this phenomenon.


\section{Conclusion}
\label{sec:conclusion}

Multi-component problems appear frequently in real-world applications and the TTP (combining the TSP and KP) has been introduced as a benchmark problem to study such problem in greater depth. Understanding the interaction of the two components is still a challenging task and we focused in this paper on the weighted TSP part of the problem. We have carried out a structural comparison of TSP variants called \wttps and \nwdtsps where the weight on nodes determined by a collection of items plays a crucial role in determining an optimal tour when the to be collected set of items is fixed.
Our results show that \wttps is closer to the TSP than the \nwdtsps and that using the fitness function of \wttps can surprisingly lead to better results when the goal is to optimise \nwdtsps.

Future work will investigate the similarity of high quality solutions of \wttps and \nwdtsps. Furthermore, evolving instances that show a significant performance difference for optimised tours of \wttps and \nwdtsps and their characterization in terms of problem features would help to push forward the understanding of the these problems.


\section*{Acknowledgment}
This work has been supported by the Australian Research Council (ARC) through grant DP160102401 and by the South Australian Government through the Research Consortium "Unlocking Complex Resources through Lean Processing".

\bibliographystyle{splncs04}
\bibliography{bib}

\end{document}